\documentclass[10pt, a4paper]{article}
\usepackage{lrec}
\usepackage{graphicx}
\usepackage{soul}

\usepackage{epstopdf}
\usepackage[utf8]{inputenc}
\usepackage{hyphenat}
\usepackage{hyperref}
\usepackage{xstring}
\usepackage{makecell}
\usepackage{multirow}

\title{A Joint Approach to Compound Splitting and Idiomatic Compound Detection}

\name{Irina Krotova$^1$, Sergey Aksenov$^2$, Ekaterina Artemova$^3$}

\address{
$^1$ MobileTeleSystems \\ 
$^2$ Sberbank of Russia\\ 
$^3$ National Research University Higher School of Economics \\
        Moscow, Russia \\
     \href{mailto:ivkrotov@mts.ru}{ivkrotov@mts.ru}, \href{mailto: aksenov.s.an@sberbank.ru}{aksenov.s.an@sberbank.ru}, \href{mailto: echernyak@hse.ru}{eartemova@hse.ru} }

\abstract{
Applications such as machine translation, speech recognition, and information retrieval require efficient handling of noun compounds as they are one of the possible sources for out-of-vocabulary (OOV) words. In-depth processing of noun compounds requires not only splitting them into smaller components (or even roots) but also the identification of instances that should remain unsplitted as they are of idiomatic nature. We develop a two-fold deep learning-based approach of noun compound splitting and idiomatic compound detection for the German language that we train using a newly collected corpus of annotated German compounds. Our neural noun compound splitter operates on a sub-word level and outperforms the current state of the art by about 5\%. \\ \newline  }

\begin{document}

\maketitleabstract

\section{Introduction} 

Compounding is a common word-formation process in Germanic languages (e.g. German, Dutch, Swedish) that poses challenges for many natural language processing applications, such as machine translation \cite{daiber2018splitting,fritzinger2010how,popovic2006morpho}, speech recognition \cite{larson2000compound}, information retrieval \cite{alfonseca2008decompounding,monz2001shallow}  and coreference resolution \cite{tuggener2016incremental}.

Difficulties are primarily caused by high productivity and low corpus frequency of compounds, which increases the vocabulary size and leads to sparse data problems. According to \cite{baroni2002predicting}, almost half (47\%) of the word types in a 28-million German newswire corpus are compounds. At the same time, 83\% of them are not frequent words or productively formed hapax legomena and have a corpus frequency of 5 or lower.

German compounds are not orthographically separated by hyphen or whitespaces and are mostly written as a single word.  For example, the equivalent of the German word \textit{Arbeitstag} is written in English as two-word compound ``working day''.

This leads to more out-of-vocabulary (OOV) words, which can not be listed in a lexicon and translated, but at the same time, may be split and represented as at least two components or roots. The right-most component is a noun, the head of the compound. The leftmost component is the modifier and can be a noun, verb, adjective, number, or a preposition.

The decomposition of a complex compound or compound splitting is a well defined, but yet not a simple task.  Two parts of the compound are not always concatenated as in \textit{Tischtennis} (``table tennis'') or \textit{Wolkenkratzer} (``skyscraper''). Compound parts can undergo morphological modifications from the normal form, such as addition (\textit{Arbeitszeit} (``working time'')) or truncation of letters (\textit{Kirchhof} (``church garden'')), umlaut (\textit{Bücherregal} (``bookshelf'')) or a combination of modifications.  All these morphological changes need to be considered to correctly split a compound into two~lemmas.

The most common way to preprocess German compounds is to split them into components before training and translation \cite{stymne2008german}. In the majority of cases, noun+noun compound nouns are realized by a determiner or adpositional phrase following the head of the compound (\textit{Haustür}~-~\textit{Tür des Hauses} (``house door'', ``front door''),\textit{ Gartenschlauch} - \textit{Schlauch für den Garten} (``garden hose'', ``hosepipe''). 

From a linguistic point of view, this refers to ``Frege's principle'', the idea of formal semantics. According to this principle, the meaning of a sentence can be deducted from the meaning of its constituents \cite{kiefer2000jelenteselmelet}. This principle can be extended to lower syntax levels, such as a phrase or a word. A compound can be tackled from this perspective because it consists of two independent nouns.

It is generally recognized that certain language phenomena, such as idioms, figures of speech (metaphors), expressions that are subjects to pragmatic interpretation, can not be interpreted in a strictly compositional way \cite{downing1977creation}. The illustration of this difference is a pair of German expressions \textit{Altmaterial} and \textit{altes Material}: the compound means ``recovered material'' whereas \textit{altes Material} describes the material as being old, where the certain meaning of the word ``old'' depends on the context.

One of the most significant and detailed works on the relationship between non-literal meaning and compositionality was written by Jan G. Kooij, see \cite{kooij1968compounds}. He distinguishes between idiomatic and non-idiomatic compounds. The meaning of the idiomatic compound cannot be explained from the constituents and the structure (consider, for example, ``egghead'' and ``egg-shell''). He claims also that some non-idiomatic compounds have meaning specialization: for example, the Dutch word \textit{huisdeur }(``house-door') consists of two words, \textit{huis }(``house'') and \textit{deur }(``door''), which are two independent words with the same meaning. However the word \textit{huisdeur} does  mean not any door in the house, and rather it refers only to the front door. He also makes an important observation, that the boundary between idiomatic and non-idiomatic compounds is not a yes-no question, but a matter of degree. A similar point of view is supported by \cite{goatly1997language}, who also emphasizes the controversy of the strict separation of literal and non-literal language usage. According to this work, they are only more or less tied to conventional meaning.

In this paper, we explore computational approaches to idiomatic meaning modeling and identification. We explore only German compounds and suggest a two-fold approach to idiomatic and literal compounds identification. The first step is to split a compound into constituents. The second step is to evaluate how likely it is that the compound is of idiomatic nature. The first step is treated as a sequence labeling task performed on the sub-word level. For each sub-word, a label, which indicates whether a split should be introduced after this very sub-word, is assigned. The second step targets at detection of idiomatic compounds and operates on the embeddings of the compound and its components. The compound is considered idiomatic, if the lexical meaning of the compound cannot be composed of the lexical meaning of its components, i.e., it is impossible to derive the embedding of the compound from the embeddings of its components. For these means, we adopted a simple yet efficient approach for compositionality detection from \cite{jana2019compositionality}.  

The contributions of this paper are as follows: we propose a new German compound splitting method, based on neural sequence models. We introduce a new dataset for the task of non-literal meaning identification and establish a baseline for this task.

\section{Related work} 

{\bf German Compound Splitting}

Methods for automatic splitting of word compounds has been studied by several research groups. 
Early approaches used dictionary-based methods as a source for full morphological analysis. \cite{koehn2003empirical} use corpora statistics and present a frequency-based approach to German compound splitting. Compound parts are identified by word frequencies and different possible splits are ranked according to the geometric mean of subword frequencies. Word modifications, such as the deletion of characters and linking elements \textit{``-s''} and \textit{``-’es''} are allowed. \cite{stymne2008german} extends this algorithm by adding the 20 most frequent morphological transformations. \cite{tuggener2016incremental} relies on character $n$-grams and their distribution. \cite{weller-di-marco-2017-simple} combines this approach with linguistic heuristics and focuses on alignment. Other researchers use unsupervised approaches to compound splitting. \cite{Macherey:2011:LCS:2002472.2002644} presents a method that does not rely on any handcrafted rules for transitional elements or morphological operations. This algorithm uses a bilingual corpus and learns morphological operations from it. \cite{ziering-van-der-plas-2016-towards} do not use parallel corpora, but rather learn ``morphological operation patterns''  from lemmatized monolingual corpora.  \cite{riedl-biemann-2016-unsupervised} explore distributional semantics; the method is based on the assumption that the constituents of a compound are semantically similar and identify the valid splitting point. They utilize a distributional thesaurus and a set of ``atomic word units'' extracted from corpus data. \cite{im2016ghost} detect the semantic relation between the constituents of the compounds.

\subsection{Word Segmentation in other languages}
The problem of word segmentation has received much attention in Chinese. Since \cite{xue2003chinese} Chinese word segmentation is addressed as a character labeling task: each character of the input sequence is labeled with one of the four labels $\mathcal{L} = \{B, M, E, S\}$, which stand for character in Begin, Middle or End of the word or Single character word. \cite{xue2003chinese} uses a maximum entropy tagger to tag each character independently. This approach was extended in \cite{peng2004chinese} to the sequence modeling task, and linear conditional random fields were used to attempt it and receive state of the art results. A neural approach to Chinese segmentation mainly uses various architectures of character level recurrent neural networks \cite{cai2016neural,zhang2018neural,cai2017fast} and very deep convolutional networks \cite{sun2017gap}. Same architectures are used for dialectal Arabic segmentation \cite{samih2017neural}.

The English word formations leads to lesser importance of the word segmentation problem. However a similar problem rises when processing social media data, hashtags in particular.  As it was shown by  \cite{berardi2011isti} hashtag segmentation for TREC microblog track 2011 \cite{soboroff2012overview} improves the quality of information retrieval, while \cite{bansal2015towards} shows that hashtag segmentation improves linking of entities extracted from tweets to a knowledge base. Both \cite{berardi2011isti,bansal2015towards} use Viterbi-like algorithm for hashtag segmentation. Following the idea of scoring segmentation candidates, \cite{reuter2016segmenting} introduces other scoring functions, which include a bigram model (2GM) and a Maximum Unknown Matching (MUM), which is adjustable to unseen words. 

A similar problem may arise outside of natural language processing scope. \cite{markovtsev2018splitting} subjected source code identifiers to analysis and use LSTM-derived splitters to extract distinct identifiers from the large chunks of code.

{\bf Compositionalty Evaluation}

\cite{hatty2018fine} proposes a combined approach for automatic term identification and investigating the understandability of terms by defining fine-grained classes of termhood and framing a classification task. They selected  400 German compounds to annotate for  termhood in the domain of cooking. Next they predicted the compound classes in three steps: compound splitting, representation of compound and its components in the feauture space and applying a neural network classifier. To split compounds CharSplit \cite{tuggener2016incremental}, CompoST \cite{cap2014morphological} and the Simple Compound Splitter \cite{weller-di-marco-2017-simple} were combined. The feature description includes word embeddings, frequency and productivity of the components. The best classifier model achieved an 80\% improvement on F1-score in comparison to the best baseline model.

\cite{horbach2016corpus} presented an annotation study on a representative dataset of literal and idiomatic uses of infinitive-verb compounds in German newspaper
and journal texts. They have collected a corpus of 6,000 instances of 6 representative infinitive-verb compounds in German, that was annotated for idiomaticity by expert lexicographers. A Naive Bayes classifier uses context features to classify instances of the verb compounds as either idiomatic or literal with an accuracy of 85\%.

\section{Dataset}

We use the dataset discussed in \cite{henrich2011determining}, GermaNet v.14.0 (2019). This is a list of 82 309 split nominal compounds extracted from a German wordnet GermaNet \cite{henrich-hinrichs-2010-gernedit-graphical}. 

The format of the compound splits is one compound per line, where the compound itself, its modifier, and the head are listed. Compound splitting is supported by automatic algorithms, combined from several compound splitters. Then all automatically split compounds are manually post-corrected and enriched with relevant properties. All modifiers in the dataset are lemmatized and in the case the modifier is ambiguous, both possibilities are specified (\textit{Laufschuhe} (``running shoes''):\textit{ lauf-} (``to run'') (en) [verb] and \textit{(der) Lauf} (``run'') [noun]).

Compounds in the dataset include compounds with different properties of head and/or modifier. The dataset includes such specific compound parts like abbreviations (\textit{SIM-Karte} (``SIM card'')), affixoids (\textit{Grundfrage} (``basic question'') - \textit{grund} (``reason, cause'')(affixoid) \textit{Frage} (``question'')), foreign words (\textit{Energydrink} (``energy drink''), confixes (\textit{Milligramm} (``milligram'') - \textit{milli} (``milli'') (confix) \textit{Gramm} (``gram'')), opaque morphemes, whose meaning is not transparent without considering its etymology (\textit{Himbeere}(``raspberry''), \textit{Lebkuchen}(``gingerbread'')), proper names (\textit{Hubbleteleskop} (``Hubble telescope'')), virtual word forms, which do not exist in the isolation (\textit{Einflussnahme} (``influence''), {Fragesteller} (``questioner'')) and word groups (\textit{Nacht-und-Nebel-Aktion} (``cloak-and-dagger operation''), \textit{Pro-Kopf-Einkommen} (``per capita income'')). As a result of the variety of compound components, the task is as close as possible to real-world challenges in machine translation of compound nouns.

The amount of the unique modifiers and head in the original dataset is much lower than the number of compounds. There are 12724 unique modifiers and 9249 unique compound heads in the dataset. Almost half of modifiers (6118) and a large part of compound heads (3752) are hapax legomenon and occur only once in the dataset. 

\subsection{Data Preprocessing and Annotation}

For the task of idiomatic compounds detection, we present the dataset of idiomatic and literal uses of German compound nouns components, based on GermaNet data. Our method includes computing word embeddings for compound nouns and their components. As the performance of word embedding degrades at low-frequent words, we limited the original dataset, namely GermaNet v.14.0 (2019), by word frequencies based on data from the DWDS corpus, constructed at the Berlin-Brandenburg Academy of Sciences (BBAW) \cite{klein2010digitale}. We produced the compound frequencies list for the Reference and Newspaper Corpora 1990 through 2019 and selected the first 5000 entries for annotation.
 
After that, we added the definitions from Duden dictionary \cite{universalworterbuch2006duden} to the list of compound nouns. Since we selected the most frequent words from the GermaNet compound list, most of them had definitions in Duden dictionary \cite{universalworterbuch2006duden}.
 
For many classification tasks, such as word sense disambiguation or named-entity recognition, there is general agreement on a standard set of categories. For the compound-related tasks, on the other hand, although numerous annotation schemes have been proposed, yet there is still little agreement and no standard categories.
 
Our annotation scheme was designed based on the principle of compositionality, described above. From this perspective, it is possible to give a compound definition using its constituents only if a compound is non-idiomatic. If a compound is not idiomatic and can not be literally translated using its constituents after splitting, the definition does not contain compound parts. See Table~\ref{tab:ann1} for examples of compounds and their definitions. According to the proposed annotation scheme  the compound \textit{Arbeitstag} is compositional, as its definition contains both constituents and the compound \textit{Schildkröte} is not compositional. 

\begin{table}[!h]
\begin{tabular}{|p{1.5cm}|p{2.8cm}|p{2.8cm}|} \hline
\textit{Arbeitstag} & \textbf{Tag}, an dem [berufliche] \textbf{Arbeit} geleistet wird oder zu leisten ist. &  Working day: the day, on which the [professional] work is done or needs to be done  \\ \hline

\textit{Schildkröte} & (besonders in Tropen und Subtropen) auf dem Land und im Wasser lebendes, sich an Land sehr schwerfällig bewegendes Tier mit Bauch- und Rückenpanzer, in den Kopf, Beine und Schwanz eingezogen werden können. & Turtle: (particularly in the tropics and subtropics) animal, which lives on land or in the water, moves slowly on land and has a shell, where its head, legs and tail can be retracted into. \\ \hline
\end{tabular}
\caption{Examples of German compounds and definitions. The constituents of compounds are bolded.}
\label{tab:ann1}
\end{table}

On the one hand, the proposed principle allows us to automatically annotate data according to the definitions from Duden dictionary \cite{universalworterbuch2006duden}. On the other hand, it makes the manual annotation task less challenging, because it is easier to distinguish between idiomatic and literal meaning of each constituent, than of the whole compound.
 
All compounds\footnote{The dataset is available:  \url{https://github.com/PragmaticsLab/kompositionsfreudigkeit}} were automatically annotated and manually post-corrected according to following schema:
\begin{itemize}
    \item[0]: both of the components can be used in the compound definition, non-idiomatic compound.
    \item[1]: the first component is idiomatic; the second is non-idiomatic.
    \item[2]: the first component is non-idiomatic; the second is idiomatic.
    \item[3]: both components are idiomatic.
    
\end{itemize}
Each compound is annotated with a value ranging from 0 to 3, which stands for the category of compound so that 0 means that compound is non-idiomatic and 3 means that the compound is idiomatic. Categories 1 and 2 can be considered borderline and partially idiomatic. The sample of the annotated dataset can be found in Table~\ref{tab:data1}.

\begin{table}[!h]
\begin{tabular}{|l|p{2cm}|p{1.5cm}|l|p{0.7cm}|} \hline
{\bf Freq} & {\bf Compound }    & {\bf Modifier } & {\bf Head} & {\bf Cate\-gory} \\ \hline
65883     & \textit{Jahrhundert}  (``century'')  &  \textit{Jahr} & \textit{Hundert}   &  0  \\  \hline
171137    & \textit{Freitag}   (``friday'')   &  \textit{frei} & \textit{Tag}   &    1   \\ \hline
33681     & \textit{Zeitpunkt} (``time moment'')   & \textit{Zeit}  &\textit{Punkt}    &     2  \\ \hline
13519     & \textit{Lebensmittel} (``foods'')  & \textit{Leben} & \textit{Mittel}      &   3    \\ \hline
\end{tabular}
\caption{Examples of annotated compounds}
\label{tab:data1}
\end{table}

\section{Problem Formulation and Models} 

\subsection{Compound Splitting Baselines}

As a baseline splitters we adopted CharSplit \cite{tuggener2016incremental}  and \texttt{SECOS}  \cite{riedl-biemann-2016-unsupervised} along with open source reference implementation of both splitters. 
CharSplit calculates the probabilities of n-grams to occur at the word’s beginning,  and middle and calculates a splitting score at each position in a compound word. 
\texttt{SECOS} leverages information from the distributional thesaurus to rank possible candidate splits. 

\subsection{Compound Splitting Models}

\begin{figure}[!h]
    \centering
    \includegraphics[width = 0.4\textwidth]{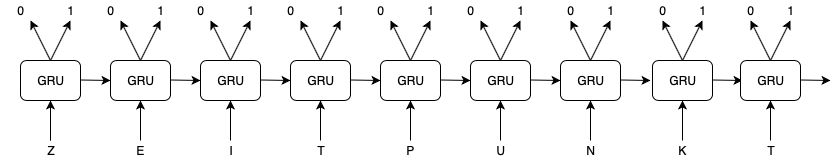}
    \caption{Our architecture}
    \label{fig:fig1}
\end{figure}

Compound splitting is treated as a sequence labeling task. We develop a set of RNN-derived models, which leverage different types of input representations and hidden units. For an architecture overview, see Figure~\ref{fig:fig1}. 

Each model is a binary classifier based on a sub-word level bidirectional recurrent neural network. The classifier determines for each sub-word, whether it is in the split position or not. Each subword is assigned with either $0$ (there is no split after this sub-word), or $1$ (there is a split after this sub-word).  The concatenation of the hidden states of the forward and backward RNN forms a feature vector for each character that is then fed to a fully connected layer. The fully connected layer has a {\it softmax} activation function that computes the probability of a split for each sub-word.

We consider the following design choices:
\begin{enumerate}
    \item {\bf sub-word definition}: a sub-word can be either a character, or a BPE sub-word unit \cite{heinzerling2018bpemb}. 
    \item {\bf RNN architecture}: we compare vanilla RNN to GRU and LSTM \cite{hochreiter1997long} architectures (we used keras\footnote{http://keras.io} implementation of each architecture)
    \item {\bf whether the embeddings are trainable\footnote{this option corresponds to the   \texttt{trainable} argument of the \texttt{Embedding} layer}}: the character embeddings are initialized randomly and thus are always learned as model parametres. The adopted pretrained BPE embeddings can be either learned as model parametrs or can be kept non-trainable and thus remain unchanged. 
\end{enumerate}

BPE tokenization has become a de-facto standard way for processing sub-words in the era of BERT \cite{devlin2019bert} and BERT-like models. Thus we decided to draw a comparison between BPE tokenization and simpler character-level models, frequently used for segmentation in Chinese \cite{xue2003chinese} or Arabic \cite{samih2017neural}. These models process input words in a character by character way so that each character is treated as a single sub-word.

The size of BPE vocabulary is one of the architecture choices. We choose from vocabulary size equal to $10^3$ and $10^4$. We choose RNN, GRU and LSTM units to be 256-dimensional. All models were trained for 30 epochs with Adam optimizer with the default learning rate equal to~$10^{-3}$. 

\subsection{Idiomatic Compounds Detection}
We establish new baselines for the task idiomatic compound detection by adopting methods of compositionality detection \cite{jana2019compositionality}.

Idiomatic compounds detection is considered as a binary classification task, where one class stands for non-idiomatic compounds (labeled with 0) and the other -- for borderline idiomatic compounds (labels 1, 2, and 3). We do not distinguish between different degrees of idiomaticity and consider a compound to be either idiomatic or not. We simply train various supervised machine learning methods on vector representations of a compound and its components. We use the following classification algorithms: logistic regression (LogReg) and gradient boosting (XGBoost). For feature representation, we use a concatenation of a compound embedding with embeddings of compound components acquired from various distributional semantics models (DSMs), such as \texttt{word2vec} \cite{mikolov2013distributed} or \texttt{fastText} \cite{joulin2017bag}.  To obtain the components of a compound, we use either the source gold standard split or our own splitter, based on Char-GRU, as it significantly outperforms other splitters.

We used two pre-trained DSMs: 
\begin{itemize}
    \item \texttt{word2vec} model pre-trained on Wikipedia. We use the Word2Vec Skip-gram model with a window size of 5 and a minimum word frequency of 10 to generate a 300-dimensional vector for each word. 
    \item \texttt{fastText} model pre-trained on Wikipedia. Similarly to \texttt{word2vec} model, the vector for each word has 300 dimensions.
\end{itemize}
The feature vector for each compound has 900 dimensions. The core difference between pre-trained DSMs is based on the way OOV words are treated.  While \texttt{word2vec} suggests using a special embedding for  unknown words (\texttt{[unk]}), \texttt{fastText} us capable to infer an embedding for any word based on n-grams.  

To detect whether the compound word is idiomatic or not, we used two classification algorithms:
\begin{itemize}
    \item Logistic Regression from \texttt{scikit-learn}\footnote{\url{https://scikit-learn.org/stable/}} with regularization strength parameter \texttt{C}=1.
    \item Gradient Boosting (XGBoost\footnote{\url{https://xgboost.readthedocs.io}}) over decision trees with 200 estimators. Minimum size of a leaf in each tree is 25. Weights for classes \texttt{0} and \texttt{1} are 1 and 10 respectively.
\end{itemize}{}

\section{Results and discussion} 

\subsection{Compound Splitting}
The results of the compound splitting experiment are presented in Table~\ref{tab:res1}. Mean accuracy values with standard deviation for 30 runs for each model are reported. 
It can be seen that all character-level models perform better than any of the BPE-level models. Character-level models learn orthographic patterns only, as they are not provided with any semantic input. Hence they are better aimed for constituent boundary detection.   
There is no significant difference, whether the BPE-embeddings are trainable or not. However, the size of BPE vocabulary matters: when trained with a larger vocabulary, the model performs better though it does not make sense to use even larger BPE vocabulary since it would include whole compounds as a single token.
Vanilla RNN architectures are always outperformed by LSTM and GRU. 

\begin{table}[!h]
\begin{center}
\begin{tabular}{|p{2cm}|p{2cm}|p{2cm}|}
      \hline
      \textbf{Model}&\textbf{Embedding layer}&\textbf{Accuracy}\\
      
      \hline
      CharSplit (Baseline)& & 0.879 \\ 
      \texttt{SECOS} (Baseline)&  & 0.914 \\ 
      
      \hline
    \hline
      \multicolumn{3}{|c|}{Char-level models } \\ \hline
    vanilla RNN & trainable & 0.915  $\pm$ 0.002\\
      \hline
 GRU & trainable & \textbf{0.956} $\pm$ 0.002\\
      \hline
biLSTM & trainable & 0.944 $\pm$ 0.003\\
      
      \hline
      \hline
      \multicolumn{3}{|c|}{BPE-level models} \\ \hline
      \multicolumn{3}{|c|}{BPE vocab size = $10^3$ } \\ \hline
vanilla RNN & non-trainable & 0.726 $\pm$ 0.003\\
      \hline
 GRU & non-trainable & 0.746 $\pm$ 0.003\\
      \hline
 biLSTM& non-trainable & 0.734 $\pm$ 0.005\\
      
            \hline
      \hline
vanilla RNN &  trainable & 0.731 $\pm$ 0.004\\
      \hline
 GRU & trainable & 0.759 $\pm$ 0.003\\
      \hline
 biLSTM & trainable & 0.752 $\pm$ 0.004\\
      
      \hline
      \hline
      \multicolumn{3}{|c|}{BPE vocab size = $10^4$ } \\ \hline

vanilla RNN & non-trainable & 0.788 $\pm$ 0.004\\
      \hline
 GRU & non-trainable & 0.802 $\pm$ 0.002\\
      \hline
biLSTM & non-trainable & 0.810 $\pm$ 0.004\\
      
      \hline
      \hline
vanilla RNN &  trainable & 0.779 $\pm$ 0.004\\
      \hline
 GRU & trainable & 0.823 $\pm$ 0.003\\
      \hline
 biLSTM & trainable & 0.825 $\pm$ 0.005\\
      
      \hline

\end{tabular}
\caption{Compound splitters performance}
\label{tab:res1}
\end{center}
\end{table}

\subsection{Idiomatic Compounds Detection}

The results of the idiomatic compound detection experiment are presented in Table~\ref{tab:res2}.
A simple model that always predicts that the word is idiomatic is referred to as {\it Dummy model}. It can be seen that classification models with pre-trained word embeddings perform significantly better than the {\it Dummy model}. 

We used two compound splitters for the task. First, we used the gold standard split from GermaNet. Second, we use our own splitter, which, according to previous experiments, happens to outperform other well-known splitters. 

Among two DSMs under consideration, \texttt{fastText}, seems to be a better source for word embeddings. As \texttt{fastText} model is capable of inferring a word embedding for out of vocabulary words,  it is less sensitive to splitter errors. 

XGBoost and logistic regression perform almost the same, with XGBoost producing slightly higher scores. Due to the high complexity of the task, the results of both classifiers are moderate. Though when compared to the {\it Dummy model}, we can stay that the classifiers are capable of detecting idiomatic compounds, which means that the task itself is cab be approached by means of machine learning and distributional semantics. 

\begin{table}[!h]
\begin{center}
\begin{tabular}{|l|l|l|}

      \hline
      \textbf{Model}&\textbf{$F_{1}$-score}\\
      \hline
      {\it Dummy model} & 0.21 \\
      \hline
      \hline
      Gold Split + \texttt{word2vec} + XGBoost & 0.567\\ 
      \hline
      Gold Split + \texttt{word2vec} + LogReg & 0.579\\
      \hline
      Gold Split + \texttt{fastText} + XGBoost & \textbf{0.584}\\
      \hline
      Gold Split + \texttt{fastText} + LogReg & 0.577\\
      \hline
      \hline

      Char-GRU Split  + \texttt{word2vec} + XGBoost & 0.545\\
      \hline
      Char-GRU Split  + \texttt{word2vec} + LogReg & 0.521\\
      \hline
      Char-GRU Split  + \texttt{fastText} + XGBoost & \textbf{0.554}\\
      \hline
      Char-GRU Split  + \texttt{fastText} + LogReg & 0.541\\
      
      \hline

\end{tabular}
\caption{Performance of idiomatic compounds detection}
\label{tab:res2}
\end{center}

\end{table}

\section{Error Analysis}
\subsection{Compound splitting}
In order to understand the errors of methods we compared, we analyzed the compounds that have been split incorrectly.

CharSplit often fails when encountering the linking element like the \textit{Fugen-s} or plural marker \textit{-(e)n-} (\textit{Gruppe-nerste} instead of \textit{Gruppen-erste} (``top of the group''), \textit{Name-nsgebung} instead of \textit{Namens-gebung} (``naming''). They are often attached to the head component of the compound noun. The second problem is splitting compounds with frequent suffixes: suffixes like \textit{'-ung'} or \textit{'-schaft'} are often recognized as a head noun (\textit{Grenzverschieb-ung} instead of \textit{Grenz-verschiebung} (``shifting of boundaries'')).

\texttt{SECOS} works in a different way and returns all the possible splitting boundaries of the compound (like \textit{Bundes-finanz-ministerium} (``Federal Ministry of Finance'')). The most frequent errors (55\% of all errors) are wrong splits of the compounds, where the modifier or both parts are compounds too (e.g. \textit{Todeszeit-punkt} (``time of death'') instead of \textit{Todes-zeitpunkt}, \textit{Arbeitszeit-raum} (``working period'') instead of \textit{Arbeits-zeitraum}, \textit{Süßwasserzier-fisch} (``freshwater'') instead of \textit{Süßwasser-zierfisch}).

The second most frequent type of \texttt{SECOS} errors are those compounds, where the modifier starts with a character or pair of characters (``er'', ``s'', ``en''), which are often used as a transitional element by compounds building (e.g. \textit{Trags-chrauber} (``autogyro'') instead of \textit{Trag-schrauber}, \textit{Gasten-gagement} (``guest engagement) instead of \textit{Gast-engagement}, \textit{Norden-gland} (``nothern England'') instead of \textit{Nord-england}).

More than one third (36\%) of the remaining errors are such compounds where at least one of the subword roots has Latin, Greek or English origin (e.g. \textit{Nitrogly-zerin} (``nitroglycerin'') instead of \textit{Nitro-glyzerin}, \textit{Lymp-hödem} (``lymphedema'') instead of \textit{Lymph-ödem}). Most of these words are scientific terms or English loan words. These roots are not frequent compared to compound parts of German origin and are not widely represented in the GermaNet data. See Table~\ref{tab:freq1} for examples of some compound parts of different origins and their absolute frequencies.

\begin{table}[!h]
\begin{center}
\begin{tabular}{|l|c|c|} \hline
{\bf Component} & {\bf Absolute frequency} \\ \hline
\textit{Tag} (``day'') & 311 \\ \hline
\textit{Land} (``country'')&  552 \\ \hline
\textit{Sport} (``sport'') &  297 \\ \hline
\textit{Lymph} (``lymphe'') & 8 \\ \hline
\textit{Ödem} (``edema'') & 4 \\ \hline
\textit{Nitro} (``nitro'') & 4 \\ \hline
\textit{Glyzerin} (``glycerin'') & 1 \\ \hline
\end{tabular}
\caption{Examples of compound parts and their absolute frequencies}
\label{tab:freq1}
\end{center}
\end{table}

\subsection{Idiomatic Compounds Detection}
The majority of the errors (413 of 477) are the non-idiomatic compounds labeled as idiomatic. The most frequent compound components of the wrong classified words can be found in Table~\ref{tab:err1}.

\begin{table}[!h]
\begin{center}
\begin{tabular}{|l|c|c|} \hline
\textit{Bund}  &  modifier   & 15 \\ \hline
\textit{Regierung}  &  modifier   &  9 \\ \hline
\textit{Staat}  &  modifier  &  8 \\ \hline
\textit{Wirtschaft}  &  modifier   & 7 \\ \hline
\textit{groß} &   modifier  &  6 \\ \hline
\textit{Wahl} &   modifier  &  6 \\ \hline
\textit{Chef} &   head &   5 \\ \hline
\textit{Verband} &  head   & 5 \\ \hline
\textit{Rat} &   head &   5 \\ \hline
\end{tabular}
\caption{Example of erroneous idiomatic compounds Detection}
\label{tab:err1}
\end{center}
\end{table}
These components are inactive metaphors, which idiomatic meaning is difficult to distinguish because of its frequency. Most of them are from the domains of politics, economics, and law on a daily basis. Most likely these compounds are challenging even for human annotators. For example, compounds with modifiers \textit{Bund-} (``national'', ``federal''), \textit{Regierung} (``government'') and \textit{Staat} (``state'', ``country'') were labeled like idiomatic.

\section{Conclusion}

In this paper, we present a two-stage approach to compound splitting and idiomatic compound detection in German. 
Our neural compound splitter is based on character-level recurrent neural networks. We outperform two well-known methods, CharSplit, and \texttt{SECOS}. To detect compounds, which should not be split, as they are of idiomatic nature, we exploit a common technique for compositionality detection.  

The suggested approach to idiomatic compound detection in its present state presents more of a proof of concept nature. It clearnly benefits from the semantic information encoded in the word embeddings though there is enough space for improvement.  One of the possible directions of the future work is to use other word embedding models, that encode not only distributional but also structural features, such as Poincare embeddings, or contextual embedding models, such as ELMo or BERT.

\section{Acknowledgements}

Ekaterina Artemova was supported by the framework of the HSE University Basic Research Program and Russian Academic  Excellence Project ``5-100''.



\section{Bibliographical References}
\label{main:ref}

\bibliographystyle{lrec}
\bibliography{lrec2020W-xample}

\begin{thebibliography}{}

\bibitem[\protect\citename{Alfonseca \bgroup et al.\egroup
  }2008]{alfonseca2008decompounding}
Alfonseca, E., Bilac, S., and Pharies, S.
\newblock (2008).
\newblock Decompounding query keywords from compounding languages.
\newblock In {\em Proceedings of the 46th Annual Meeting of the Association for
  Computational Linguistics on Human Language Technologies: Short Papers},
  HLT-Short '08, pages 253--256, Stroudsburg, PA, USA. Association for
  Computational Linguistics.

\bibitem[\protect\citename{Bansal \bgroup et al.\egroup
  }2015]{bansal2015towards}
Bansal, P., Jain, S., and Varma, V.
\newblock (2015).
\newblock Towards semantic retrieval of hashtags in microblogs.
\newblock In {\em Proceedings of the 24th International Conference on World
  Wide Web}, pages 7--8.

\bibitem[\protect\citename{Baroni \bgroup et al.\egroup
  }2002]{baroni2002predicting}
Baroni, M., Matiasek, J., and Trost, H.
\newblock (2002).
\newblock Predicting the components of german nominal compounds.
\newblock In {\em ECAI}, pages 470--474.

\bibitem[\protect\citename{Berardi \bgroup et al.\egroup
  }2011]{berardi2011isti}
Berardi, G., Esuli, A., Marcheggiani, D., and Sebastiani, F.
\newblock (2011).
\newblock Isti@ trec microblog track 2011: Exploring the use of hashtag
  segmentation and text quality ranking.
\newblock In {\em TREC}.

\bibitem[\protect\citename{Cai and Zhao}2016]{cai2016neural}
Cai, D. and Zhao, H.
\newblock (2016).
\newblock Neural word segmentation learning for chinese.
\newblock In {\em Proceedings of the 54th Annual Meeting of the Association for
  Computational Linguistics (Volume 1: Long Papers)}, pages 409--420.

\bibitem[\protect\citename{Cai \bgroup et al.\egroup }2017]{cai2017fast}
Cai, D., Zhao, H., Zhang, Z., Xin, Y., Wu, Y., and Huang, F.
\newblock (2017).
\newblock Fast and accurate neural word segmentation for chinese.
\newblock In {\em Proceedings of the 55th Annual Meeting of the Association for
  Computational Linguistics (Volume 2: Short Papers)}, pages 608--615.

\bibitem[\protect\citename{Cap}2014]{cap2014morphological}
Cap, F.
\newblock (2014).
\newblock Morphological processing of compounds for statistical machine
  translation.

\bibitem[\protect\citename{Daiber \bgroup et al.\egroup
  }2015]{daiber2018splitting}
Daiber, J., Quiroz, L., Wechsler, R., and Frank, S.
\newblock (2015).
\newblock Splitting compounds by semantic analogy.
\newblock {\em CoRR}, abs/1509.04473.

\bibitem[\protect\citename{Devlin \bgroup et al.\egroup }2019]{devlin2019bert}
Devlin, J., Chang, M.-W., Lee, K., and Toutanova, K.
\newblock (2019).
\newblock Bert: Pre-training of deep bidirectional transformers for language
  understanding.
\newblock In {\em Proceedings of the 2019 Conference of the North American
  Chapter of the Association for Computational Linguistics: Human Language
  Technologies, Volume 1 (Long and Short Papers)}, pages 4171--4186.

\bibitem[\protect\citename{Downing}1977]{downing1977creation}
Downing, P.
\newblock (1977).
\newblock On the creation and use of english compound nouns.
\newblock {\em Language}, pages 810--842.

\bibitem[\protect\citename{Duden~Universalw{\"o}rterbuch}2006]{universalworterbuch2006duden}
Duden~Universalw{\"o}rterbuch, D.
\newblock (2006).
\newblock Duden.
\newblock {\em Deutsches Universalw{\"o}rterbuch. CD-ROM. Hrsg. vd
  Dudenredaktion. Mannheim/Leipzig/Wien/Z{\"u}rich: Dudenverlag}.

\bibitem[\protect\citename{Fritzinger and Fraser}2010]{fritzinger2010how}
Fritzinger, F. and Fraser, A.
\newblock (2010).
\newblock How to avoid burning ducks: Combining linguistic analysis and corpus
  statistics for german compound processing.
\newblock In {\em Proceedings of the Joint Fifth Workshop on Statistical
  Machine Translation and MetricsMATR}, WMT '10, pages 224--234, Stroudsburg,
  PA, USA. Association for Computational Linguistics.

\bibitem[\protect\citename{Goatly}1997]{goatly1997language}
Goatly, A.
\newblock (1997).
\newblock {\em The language of metaphors}.
\newblock Routledge.

\bibitem[\protect\citename{H{\"a}tty and im Walde}2018]{hatty2018fine}
H{\"a}tty, A. and im~Walde, S.~S.
\newblock (2018).
\newblock Fine-grained termhood prediction for german compound terms using
  neural networks.
\newblock In {\em Proceedings of the Joint Workshop on Linguistic Annotation,
  Multiword Expressions and Constructions (LAW-MWE-CxG-2018)}, pages 62--73.

\bibitem[\protect\citename{Heinzerling and Strube}2018]{heinzerling2018bpemb}
Heinzerling, B. and Strube, M.
\newblock (2018).
\newblock Bpemb: Tokenization-free pre-trained subword embeddings in 275
  languages.
\newblock In {\em Proceedings of the Eleventh International Conference on
  Language Resources and Evaluation (LREC-2018)}.

\bibitem[\protect\citename{Henrich and
  Hinrichs}2010]{henrich-hinrichs-2010-gernedit-graphical}
Henrich, V. and Hinrichs, E.
\newblock (2010).
\newblock {G}ern{E}di{T}: A graphical tool for {G}erma{N}et development.
\newblock In {\em Proceedings of the {ACL} 2010 System Demonstrations}, pages
  19--24, Uppsala, Sweden, July. Association for Computational Linguistics.

\bibitem[\protect\citename{Henrich and Hinrichs}2011]{henrich2011determining}
Henrich, V. and Hinrichs, E.
\newblock (2011).
\newblock Determining immediate constituents of compounds in germanet.
\newblock In {\em Proceedings of the international conference recent advances
  in natural language processing 2011}, pages 420--426.

\bibitem[\protect\citename{Hochreiter and Schmidhuber}1997]{hochreiter1997long}
Hochreiter, S. and Schmidhuber, J.
\newblock (1997).
\newblock Long short-term memory.
\newblock {\em Neural computation}, 9(8):1735--1780.

\bibitem[\protect\citename{Horbach \bgroup et al.\egroup
  }2016]{horbach2016corpus}
Horbach, A., Hensler, A., Krome, S., Prange, J., Scholze-Stubenrecht, W.,
  Steffen, D., Thater, S., Wellner, C., and Pinkal, M.
\newblock (2016).
\newblock A corpus of literal and idiomatic uses of german infinitive-verb
  compounds.
\newblock In {\em Proceedings of the Tenth International Conference on Language
  Resources and Evaluation (LREC'16)}, pages 836--841.

\bibitem[\protect\citename{Jana \bgroup et al.\egroup
  }2019]{jana2019compositionality}
Jana, A., Puzyrev, D., Panchenko, A., Goyal, P., Biemann, C., and Mukherjee, A.
\newblock (2019).
\newblock On the compositionality prediction of noun phrases using poincar{\'e}
  embeddings.
\newblock In {\em Proceedings of the 57th Annual Meeting of the Association for
  Computational Linguistics}, pages 3263--3274, Florence, Italy, July.
  Association for Computational Linguistics.

\bibitem[\protect\citename{Joulin \bgroup et al.\egroup }2017]{joulin2017bag}
Joulin, A., Grave, E., and Mikolov, P. B.~T.
\newblock (2017).
\newblock Bag of tricks for efficient text classification.
\newblock {\em EACL 2017}, page 427.

\bibitem[\protect\citename{Kiefer}2000]{kiefer2000jelenteselmelet}
Kiefer, F.
\newblock (2000).
\newblock Jelent{\'e}selm{\'e}let, corvina.

\bibitem[\protect\citename{Klein and Geyken}2010]{klein2010digitale}
Klein, W. and Geyken, A.
\newblock (2010).
\newblock Das digitale w{\"o}rterbuch der deutschen sprache (dwds).
\newblock In {\em Lexicographica: International annual for lexicography}, pages
  79--96. De Gruyter.

\bibitem[\protect\citename{Koehn}2003]{koehn2003empirical}
Koehn, P.
\newblock (2003).
\newblock Empirical methods for compound splitting.
\newblock In {\em Proceedings of EACL, 2003}, pages 187--193.

\bibitem[\protect\citename{Kooij}1968]{kooij1968compounds}
Kooij, J.~G.
\newblock (1968).
\newblock Compounds and idioms.
\newblock {\em Lingua}, 21:250--268.

\bibitem[\protect\citename{Larson \bgroup et al.\egroup
  }2000]{larson2000compound}
Larson, M., Willett, D., K{\"o}hler, J., and Rigoll, G.
\newblock (2000).
\newblock Compound splitting and lexical unit recombination for improved
  performance of a speech recognition system for german parliamentary speeches.
\newblock In {\em Sixth International Conference on Spoken Language
  Processing}.

\bibitem[\protect\citename{Macherey \bgroup et al.\egroup
  }2011]{Macherey:2011:LCS:2002472.2002644}
Macherey, K., Dai, A.~M., Talbot, D., Popat, A.~C., and Och, F.
\newblock (2011).
\newblock Language-independent compound splitting with morphological
  operations.
\newblock In {\em Proceedings of the 49th Annual Meeting of the Association for
  Computational Linguistics: Human Language Technologies - Volume 1}, HLT '11,
  pages 1395--1404, Stroudsburg, PA, USA. Association for Computational
  Linguistics.

\bibitem[\protect\citename{Markovtsev \bgroup et al.\egroup
  }2018]{markovtsev2018splitting}
Markovtsev, V., Long, W., Bulychev, E., Keramitas, R., Slavnov, K., and
  Markowski, G.
\newblock (2018).
\newblock Splitting source code identifiers using bidirectional lstm recurrent
  neural network.
\newblock {\em arXiv preprint arXiv:1805.11651}.

\bibitem[\protect\citename{Mikolov \bgroup et al.\egroup
  }2013]{mikolov2013distributed}
Mikolov, T., Sutskever, I., Chen, K., Corrado, G.~S., and Dean, J.
\newblock (2013).
\newblock Distributed representations of words and phrases and their
  compositionality.
\newblock In {\em Advances in neural information processing systems}, pages
  3111--3119.

\bibitem[\protect\citename{Monz and De~Rijke}2001]{monz2001shallow}
Monz, C. and De~Rijke, M.
\newblock (2001).
\newblock Shallow morphological analysis in monolingual information retrieval
  for dutch, german, and italian.
\newblock In {\em Workshop of the Cross-Language Evaluation Forum for European
  Languages}, pages 262--277. Springer.

\bibitem[\protect\citename{Peng \bgroup et al.\egroup }2004]{peng2004chinese}
Peng, F., Feng, F., and McCallum, A.
\newblock (2004).
\newblock Chinese segmentation and new word detection using conditional random
  fields.
\newblock In {\em Proceedings of the 20th international conference on
  Computational Linguistics}, page 562. Association for Computational
  Linguistics.

\bibitem[\protect\citename{Popovi{\'c} \bgroup et al.\egroup
  }2006]{popovic2006morpho}
Popovi{\'c}, M., de~Gispert, A., Gupta, D., Lambert, P., Ney, H., Mari{\~n}o,
  J.~B., Federico, M., and Banchs, R.
\newblock (2006).
\newblock Morpho-syntactic information for automatic error analysis of
  statistical machine translation output.
\newblock In {\em Proceedings on the Workshop on Statistical Machine
  Translation}, pages 1--6, New York City, June. Association for Computational
  Linguistics.

\bibitem[\protect\citename{Reuter \bgroup et al.\egroup
  }2016]{reuter2016segmenting}
Reuter, J., Pereira-Martins, J., and Kalita, J.
\newblock (2016).
\newblock Segmenting twitter hashtags.
\newblock {\em Intl. J. on Natural Lang. Computing}, 5(4).

\bibitem[\protect\citename{Riedl and
  Biemann}2016]{riedl-biemann-2016-unsupervised}
Riedl, M. and Biemann, C.
\newblock (2016).
\newblock Unsupervised compound splitting with distributional semantics rivals
  supervised methods.
\newblock In {\em Proceedings of the 2016 Conference of the North {A}merican
  Chapter of the Association for Computational Linguistics: Human Language
  Technologies}, pages 617--622, San Diego, California, June. Association for
  Computational Linguistics.

\bibitem[\protect\citename{Samih \bgroup et al.\egroup }2017]{samih2017neural}
Samih, Y., Attia, M., Eldesouki, M., Abdelali, A., Mubarak, H., Kallmeyer, L.,
  and Darwish, K.
\newblock (2017).
\newblock A neural architecture for dialectal arabic segmentation.
\newblock In {\em Proceedings of the Third Arabic Natural Language Processing
  Workshop}, pages 46--54.

\bibitem[\protect\citename{Schulte~im Walde \bgroup et al.\egroup
  }2016]{im2016ghost}
Schulte~im Walde, S.~S., H{\"a}tty, A., Bott, S., and Khvtisavrishvili, N.
\newblock (2016).
\newblock Ghost-nn: A representative gold standard of german noun-noun
  compounds.
\newblock In {\em Proceedings of the Tenth International Conference on Language
  Resources and Evaluation (LREC'16)}, pages 2285--2292.

\bibitem[\protect\citename{Soboroff \bgroup et al.\egroup
  }2012]{soboroff2012overview}
Soboroff, I., Ounis, I., Macdonald, C., and Lin, J.~J.
\newblock (2012).
\newblock Overview of the trec-2012 microblog track.
\newblock In {\em TREC}, volume 2012, page~20.

\bibitem[\protect\citename{Stymne}2008]{stymne2008german}
Stymne, S.
\newblock (2008).
\newblock German compounds in factored statistical machine translation.
\newblock In {\em International Conference on Natural Language Processing},
  pages 464--475. Springer.

\bibitem[\protect\citename{Sun \bgroup et al.\egroup }2017]{sun2017gap}
Sun, Z., Shen, G., and Deng, Z.
\newblock (2017).
\newblock A gap-based framework for chinese word segmentation via very deep
  convolutional networks.
\newblock {\em arXiv preprint arXiv:1712.09509}.

\bibitem[\protect\citename{Tuggener}2016]{tuggener2016incremental}
Tuggener, D.
\newblock (2016).
\newblock {\em Incremental coreference resolution for German}.
\newblock {Ph.D.} thesis, Universit{\"a}t Z{\"u}rich.

\bibitem[\protect\citename{Weller-Di~Marco}2017]{weller-di-marco-2017-simple}
Weller-Di~Marco, M.
\newblock (2017).
\newblock Simple compound splitting for {G}erman.
\newblock In {\em Proceedings of the 13th Workshop on Multiword Expressions
  ({MWE} 2017)}, pages 161--166, Valencia, Spain, April. Association for
  Computational Linguistics.

\bibitem[\protect\citename{Xue and Shen}2003]{xue2003chinese}
Xue, N. and Shen, L.
\newblock (2003).
\newblock Chinese word segmentation as lmr tagging.
\newblock In {\em Proceedings of the second SIGHAN workshop on Chinese language
  processing-Volume 17}, pages 176--179. Association for Computational
  Linguistics.

\bibitem[\protect\citename{Zhang \bgroup et al.\egroup }2018]{zhang2018neural}
Zhang, Q., Liu, X., and Fu, J.
\newblock (2018).
\newblock Neural networks incorporating dictionaries for chinese word
  segmentation.
\newblock In {\em Thirty-Second AAAI Conference on Artificial Intelligence}.

\bibitem[\protect\citename{Ziering and van~der
  Plas}2016]{ziering-van-der-plas-2016-towards}
Ziering, P. and van~der Plas, L.
\newblock (2016).
\newblock Towards unsupervised and language-independent compound splitting
  using inflectional morphological transformations.
\newblock In {\em Proceedings of the 2016 Conference of the North {A}merican
  Chapter of the Association for Computational Linguistics: Human Language
  Technologies}, pages 644--653, San Diego, California, June. Association for
  Computational Linguistics.

\end{thebibliography}



\end{document}